\pdfoutput=1
\PassOptionsToPackage{table}{xcolor}

\documentclass[11pt]{article}

\usepackage{acl}
\usepackage{times}
\usepackage{latexsym}
\usepackage{graphicx}
\usepackage{tabularx}

\usepackage{caption}
\usepackage{subcaption}
\usepackage{amsmath}
\usepackage{arydshln}

\definecolor{bostonuniversityred}{rgb}{0.8, 0.0, 0.0}
\definecolor{applegreen}{rgb}{0.45, 0.71, 0.0}

\usepackage[linguistics]{forest} 
\usepackage{algorithmic,algorithm}
\renewcommand{\algorithmiccomment}[1]{\bgroup\hfill$\triangleright$~#1\egroup}

\usepackage{arydshln}

\usepackage[T1]{fontenc}

\usepackage[utf8]{inputenc}

\usepackage{microtype}

\usepackage{inconsolata}

\usepackage[table]{xcolor}

\title{Neural Automated Writing Evaluation with Corrective Feedback}

\author{
Izia Xiaoxiao Wang$^{1*}$~~~ Xihan Wu$^{2*}$~~~ Edith Coates$^{3}$\thanks{Equally contributed authors.}~~~ Min Zeng$^{2}$~~~ Jiexin Kuang$^{2}$\\
\textbf{Siliang Liu$^{2}$}~~~ \textbf{Mengyang Qiu$^{4}$}~~~ \textbf{Jungyeul Park$^{2}$}\\
$^{1}$Linguistik Zentrum Zürich, Universität Zürich, Schweiz\\
$^{2}$Department of Linguistics, The University of British Columbia, Canada\\
$^{3}$Department of Mathematics, The University of British Columbia, Canada\\
$^{4}$Department of Psychology, Trent University, Canada\\
}

\begin{document}
\maketitle
\begin{abstract}
The utilization of technology in second language learning and teaching has become ubiquitous. For the assessment of writing specifically, automated writing evaluation (AWE) and grammatical error correction (GEC) have become immensely popular and effective methods for enhancing writing proficiency and delivering instant and individualized feedback to learners.
By leveraging the power of natural language processing (NLP) and machine learning algorithms, AWE and GEC systems have been developed separately to provide language learners with automated corrective feedback and more accurate and unbiased scoring that would otherwise be subject to examiners.
In this paper, we propose an integrated system for automated writing evaluation with corrective feedback 
as a means of bridging the gap between AWE and GEC results for second language learners.
This system enables language learners to simulate the essay writing tests: a student writes and submits an essay, and the system returns the assessment of 
the writing along with suggested grammatical error corrections. 
Given that automated scoring and grammatical correction are more efficient and cost-effective than human grading, 
this integrated system 
would also alleviate the burden of manually correcting innumerable essays.
A link to the demonstration video: \url{https://youtu.be/CoqmBqqcXio}
\end{abstract}

\section{Introduction}

Second language (L2) instructors and learners both believe that instructors are not fulfilling their duties if they fail to offer sufficient feedback on the essay submissions \citep{cho:2018}.
However, providing and revising feedback can be incredibly time-consuming and laborious tasks for both language instructors and learners. 
Automated writing evaluation and grammatical error correction tools would offer potential solutions for alleviating the workload for language instructors and
reducing the overhead of trying to understand the instructor's feedback for
language learners.
These systems help to meet the need for more efficient practices in the digital age, providing instantaneous scoring with corrective feedback.

The basic term AWE is defined as ``the process of evaluating and scoring written prose via computer programs'' \citep{shermis-burstein:2003}. 
In essence, this system predicts continuous values for a holistic score or a set of trait scores, thereby providing a comprehensive assessment of the writing quality.
AWE systems require multiple source prompt (\textit{e.g.} cross-prompt) scoring instead of prompt-specific scoring.
In practical usage, a set of scores across different rubrics is much more in need, rather than a holistic and overall writing quality assessment.
Therefore, we argue that an effective AWE system should incorporate cross-prompt rubric scoring for constructive results.

GEC is {the automation of} detecting and correcting grammatical errors in the text.
As the number of second language learners of English rises, and the demand {for} timely feedback to better facilitate their learning intensifies, GEC 
{has gained increasing popularity and attention}
in both academia and industry
in recent years.
Currently, a common approach is to treat GEC as a translation task where incorrect sentences are translated into correct sentences.

In this paper, we aim to demonstrate a system that would contribute to the integration of AWE and GEC {and} is designed to facilitate the language learning process and promote better learning outcomes. 
While there exist various {approaches} that attempt to leverage the efficiency of GEC or AWE systems to {give away} seemingly instantaneous but virtually momentary solutions for language learners, our objective is to introduce a robust and sensible approach to transform traditional language education. 
{Our approach empowers language learners to "write it right" by enabling simulated examination situations.} 

\section{Previous Work}

\paragraph{AWE} 
\citet{jacobs-EtAl:1981} first proposed the prototype for a set of components that in conjunction capture the communicative effectiveness of compositions holistically. Later, it was developed and refined by \citet{connorlinton-polio:2014}. Moreover, a variation of the same holistic evaluation on essays is implicitly used in developing the different attributes in the ASAP++ data set \citep{mathias-bhattacharyya:2018:LREC} for English.
Currently, neural models have been dominating AWE systems. \citet{ke-ng:2019:IJCAI}, \citet{ramesh-sanampudi:2021}, and \citet{uto:2021} have summarized recent neural models well.
For automatic essay scoring, there are two main model types. 
Firstly, in RNN-based models, the RNN output is sent to mean-over-time to aggregate the input to the fixed length vector and a linear layer for the scalar value \citep{taghipour-ng:2016:EMNLP} , or alternatively, a simple BiLSTM to the linear layer is used for predicting essay scores \citep{alikaniotis-alikaniotis-rei:2016:ACL}. 
Secondly, transformer-based models, \textit{e.g.}, \texttt{BERT} with BiLSTM with attention \citep{nadeem-EtAl:2019:BEA} or \texttt{BERT} concatenated with handcrafted features
\citep{uto-xie-ueno:2020:COLING}, can be used to predict the score.
Fine-tuning \texttt{BERT} using multiple losses, including regression loss and reranking loss, for constraining automated essay scores has been shown to produce state-of-the-art results \citep{yang-EtAl:2020:EMNLPFindings}.

\paragraph{GEC} Similar to AWE, neural models have also dominated GEC systems{,} where \citet{wang-EtAl:2021} presents a comprehensive review.
In general, recent state-of-the-art systems can be divided into two categories: sequence-to-sequence  NMT-based approaches and sequence tagging edit-based approaches. 
For NMT-based approaches, a transformer-based encoder-decoder architecture is typically used, where source (or incorrect) sentences are encoded into a hidden state with a multi-head self-attention layer \citep{vaswani-EtAl:2017:NIPS}.
Unlike a typical translation task, GEC only requires changing a few words in a sentence. Incorporating a copying mechanism into seq2seq (that is, copying unchanged words from original sentences) has been shown to significantly boost GEC performance \citep{zhao-EtAl:2019:NAACL}.
This unique characteristic of GEC (\textit{i.e.}, the high overlap between the source and target sentences) has also led to the development of edit-based approaches \citep{omelianchuk-etal-2020-gector}.
Edit-based models also use a transformer-based architecture. 
However, instead of predicting complete sentences, these models are trained to predict a series of editing operations (\textit{e.g.}, delete, append, and replace), which has dramatically improved the speed of inference while maintaining high performance.
Given that there is a lack of sufficient parallel training data in GEC, data augmentation via artificial error generation has been shown to boost performance in both NMT-based and edit-based approaches \citep{stahlberg-kumar-2021-synthetic}.
In addition, incorporating large pre-trained masked language models, such as \texttt{BERT} and its variants \citep{devlin-EtAl:2019:NAACL}, in pre-training and/or fine-tuning stages has yielded better results \citep{kaneko-EtAl:2020:ACL}.

{
Only a few previous work have shown for their efforts to integrate pre-neural  AWE and GEC systems together where they have focused on perceived feedback of language learners \citep{ranalli:2018,oneill-russell:2019,zhang:2020,ariyanto-mukminatien-tresnadewi:2021,reynolds-kao-huang:2021}.
Whereas empirical studies have explored students' perception, beliefs, and preferences towards feedback, {their} influential role in their writing performance has remained unknown \citep{truscott:1999,ferris:1999,chandler:2003,ferris:2004,ferris:2014}.
Hence, this paper concentrates on the integration of the AWE and GEC systems {and its impact on language education.}
The combination of neural AWE and GEC systems is introduced for the first time to achieve results that are close to the state-of-the-art (SOTA).
}

\section{System Description}

\begin{figure}[!ht]
\centering
\resizebox{.4\textwidth}{!}{    
\includegraphics{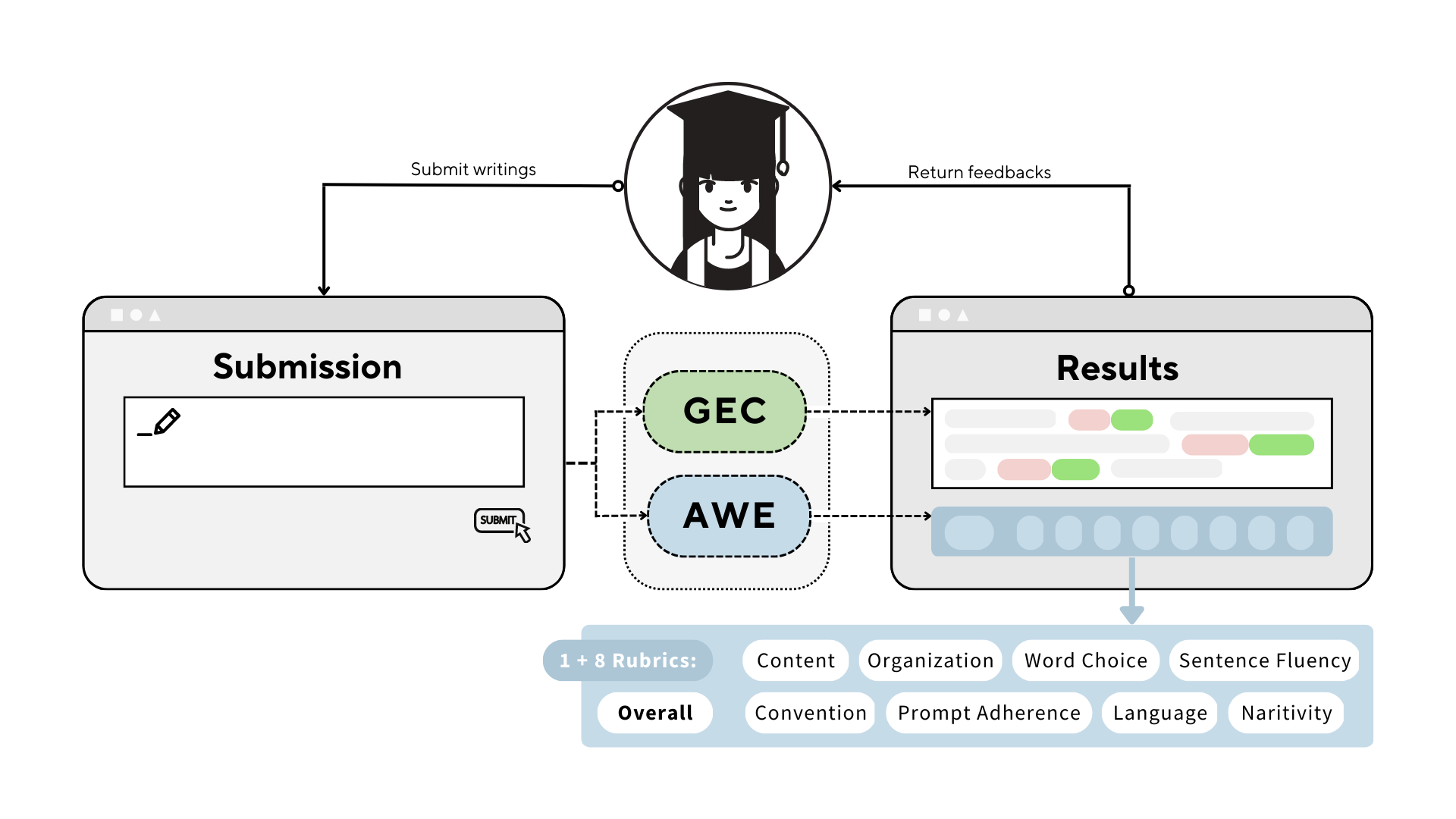}
}
\caption{System workflow of {integrated} AWE and GEC for language learners. 
A user flow can be applied to simulate examination situations{:}
{t}he language learners receive instant objective scoring results and corrective feedback.}
\label{flowchart}
\end{figure}

The flowchart in Figure~\ref{flowchart} demonstrates the learner's interactions with the system, from the initial step of submitting the writing to receiving the AWE and GEC results. 
Adapted to the language learners' perspectives, the procedures are disassembled as the following:

(i) After completing the writings, language learners submit their writings to the system. 
(ii) Taken the submitted writing as input, the system compiles both AWE and GEC components with it.
(iii) For the GEC, the output presents in-line grammatical error correction{s} with the original writing.
(iv) For the AWE, the output offers one overall and eight rubric scores, which 
{delivers the complete set of the evaluation results from the AWE system.}
(v) At the output interface, the language learner receives the combined feedback from the two integrated outputs generated from AWE and GEC.

\begin{figure}[!ht]
\centering
\resizebox{.4\textwidth}{!}{
\includegraphics{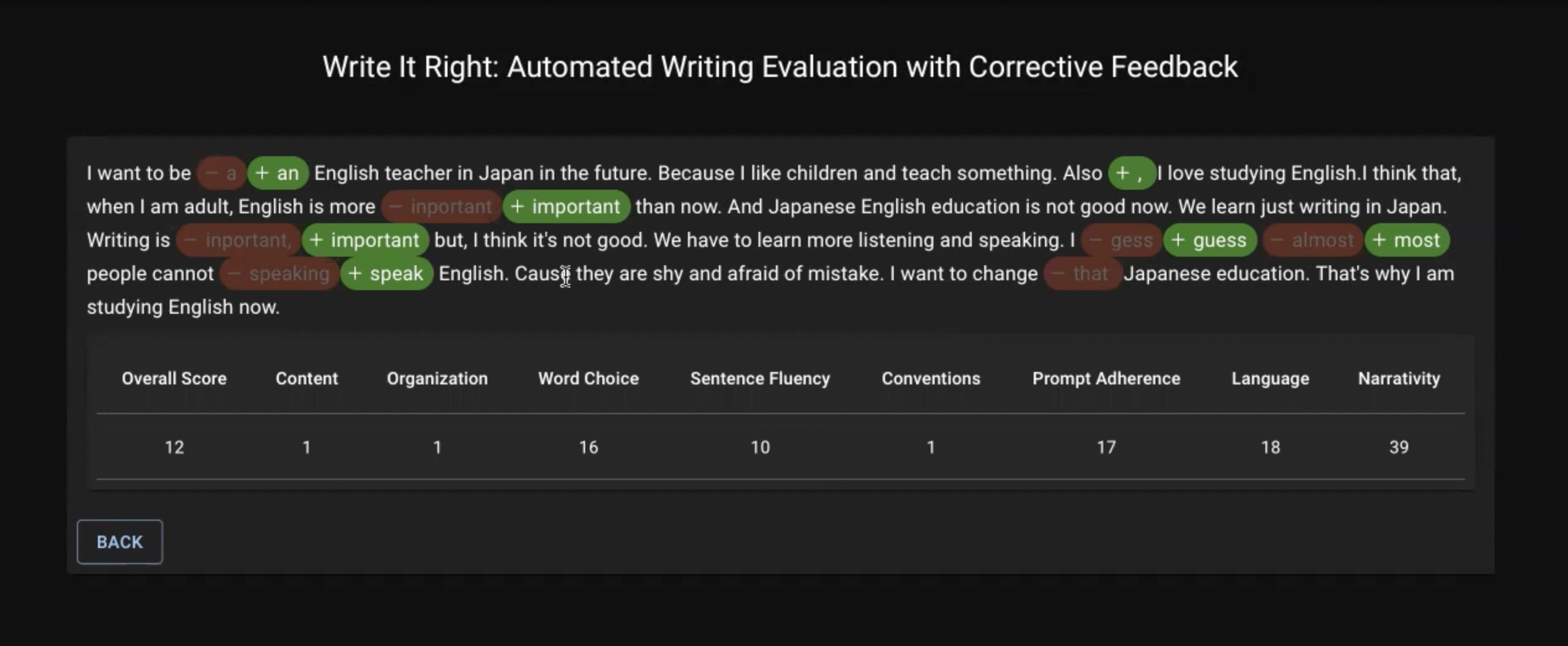}
}
\caption{{Interface screenshot} for AWE and GEC results: {tokens} in the {faded} red rectangles have been deleted; tokens in the green rectangles {are corresponding grammatical corrections} inserted by GEC{;} AWE scores are scaled to 0-100.}
\label{screenshot}
\end{figure}

\renewcommand\labelenumi{(\theenumi)}
\begin{enumerate}\itemsep-.3em 
\item\label{sent1} *\textit{I {\color{bostonuniversityred}gess} {\color{bostonuniversityred}almost} people cannot {\color{bostonuniversityred}speaking} English.}
\item\label{sent2}  {\color{white}*}\textit{I {\color{applegreen}guess} {\color{applegreen}most} people cannot {\color{applegreen}speak} English.}
\end{enumerate}

Figure~\ref{screenshot} shows a{n interface} screenshot of {the integrated} AWE and GEC results for the sentence from {a} language learner's writing in \eqref{sent1} and grammatical error{s} corrected in \eqref{sent2}. 
AWE results show the absolute values between 0 and 100 regardless of their proficiency level. {These AWE scores can also be normalized} based on the learner's proficiency level using min-max normalization if it is required.


\paragraph{AWE}
For AWE models,\footnote{Since we are working with L2 English learners, we prefer to use AWE instead of AES (automatic essay scoring), which is designated for L1 English learners.} we use Automated Student Assessment Prize (ASAP)\footnote{\url{https://www.kaggle.com/competitions/asap-aes}} and its extension ASAP++ \citep{mathias-bhattacharyya:2018:LREC} for English. 
ASAP was introduced as part of a Kaggle competition in 2012{.} {It} has since become widely used in prompt-specific \citep{alikaniotis-alikaniotis-rei:2016:ACL,dong-zhang-yang:2017:CONLL} and cross-prompt \citep{cummins-zhang-briscoe:2016:ACL,ridley-EtAl:2021:AAAI,jiang-etal-2023-improving,chen-li-2023-pmaes,do-etal-2023-prompt} automated scoring systems. 
Since only two of eight sets in ASAP have scores for individual essay attributes, ASAP++ annotates other rubric scoring for six sets of ASAP.  
The current ASAP and ASAP++ datasets contain eight different prompt essay sets with from three to  five rubric human-rated scores:
\textit{e.g.} \texttt{content}, \texttt{organization}, \texttt{word choice}, \texttt{sentence fluency}, \texttt{conventions}, \texttt{prompt adherence}, \texttt{language}, \texttt{narrativity} as well as \texttt{overall}. 
ASAP consists of only overall scores for Prompts 1-6, and different rubric scores (content, organization, conventions for Prompt 7, and content, organization, word choice, sentence fluency, conventions for Prompt 8). 
ASAP++ provides various rubric scores for Prompts 1–6 (content, organization, word choice, sentence fluency, convention for Prompts 1-2, and content, prompt adherence, language, narrativity for Prompt 3-6) to supplement the original overall scores from ASAP (Prompt 1-6).

For the current system, we use the denoised ASAP dataset, which we employ simple text denoising techniques using prompting. 
The original ASAP dataset employed a named entity recognition approach using Stanford CoreNLP \citep{manning-etal-2014-stanford} and a range of pattern matching rules to eliminate personally identifying information from the essays. Consequently, entities within the text are identified and replaced with strings initiated with the `@' symbol, such as \texttt{@PERSON1}.
Furthermore, more than 5\% of sentences exhibit encoding issues where UTF-8 symbols are not correctly displayed, in addition to the presence of non-word entities.
We classify sentences containing encoding issues and non-word entities as noise, and subsequently, undergo a denoising process to address them. Our hypothesis posits that enhancing text quality through denoising will yield improved linear regression results in AES. 
For this process, we employ two prompts with \texttt{gpt-3.5-turbo-instruct}: one to address encoding errors and another to replace non-word entities with arbitrary entity names sequentially.
It is noteworthy that \texttt{gpt-3.5-turbo-instruct} has a tendency to correct grammatical errors in the original text during sentence generation. To restore the original words, we utilize the \texttt{.m2} annotation generated by ERRANT \citep{bryant-felice-briscoe:2017:ACL}. This annotation delineates the modifications made in the original text, allowing us to retain only the symbols with corrected encoding and the replaced non-word entities.
After text denoising, we utilize \texttt{roberta-base} \citep{liu-et-al-2019-roberta} for the linear regression task. 
Detailed prompt-by-prompt results are presented in Table~\ref{results-details}, including the evaluation conducted on the original text to ensure a fair comparison (\textsc{Cleaned}).
During training, we utilize the default values of the \texttt{Trainer} class,\footnote{\url{https://huggingface.co/docs/transformers/main_classes/trainer}} employing the RoBERTa base model.
We normalize the score to the range of 0 and 1, and  we multiply the results by 100 to calculate QWK using the standard evaluation script provided by \texttt{ets.org}.

\begin{table*}[!ht]
\centering
\resizebox{\textwidth}{!}
{    
{\footnotesize
\begin{tabular}{c  | cccc cccc | c}\hline 
&  Prompt 1 & Prompt 2 & Prompt 3& Prompt 4& Prompt 5& Prompt 6& Prompt 7& Prompt 8 & \textsc{Average}\\ \hline
\textsc{Original} & 
0.6187 & 
0.6308 & 
0.6962 & 
0.6626 & 
0.7158          & 
0.5924          & 
0.4622          & 
0.4586          & 
0.6047 \\

\textsc{Cleaned} & 
0.7344 & 
0.5485 & 
0.6978 & 
0.6540 & 
0.7416 & 
0.6156 & 
0.4637 & 
0.4589 & 
0.6143 \\
\hline 
    \end{tabular}
}
}
\caption{Prompt-by-prompt QWK results: \textit{e.g.} \texttt{prompt 1} represents that prompt 1 is used as a test set, and prompts 2-7 as a training set, and prompt 8 as a dev set.}
\label{results-details}
\end{table*}


\paragraph{GEC}
We build seq2seq GEC models using \texttt{fairseq} \citep{ott-etal-2019-fairseq}. 
We explore ensemble models that incorporate different pre-trained large language models (LLMs), including \texttt{BERT} and \texttt{BART}.
The key difference between the two LLMs is in their pre-training objectives.
\texttt{BERT} is trained using a masked language model{l}ing objective, where a certain percentage of the input tokens are masked, and the model is trained to predict the original tokens. 
In contrast, \texttt{BART} is trained using a denoising auto-encoding objective, where the model is trained to reconstruct the original input text from a corrupted version of the input. 
Our {current} training setup is largely based on \citet{kaneko-EtAl:2020:ACL}. 
We use four GEC datasets -- FCE, NUCLE, W\&I +LOCNESS, and Lang-8 -- provided by the BEA 2019 Shared Task for training \citep{bryant-etal-2019-bea}.\footnote{\url{https://www.cl.cam.ac.uk/research/nl/bea2019st}} 
All the incorrect sentences are spell-checked and added back in order to augment the size of the original training set (1,157,324 sentence pairs), resulting in 1,718,693 sentence pairs in total. 
6,575 sentence pairs from the FCE and W\&I+LOCNESS development sets are used as validation.
Our best model (using \texttt{BERT}) achieves a 65.29 F$_{0.5}$ score on the BEA 2019 test set.
We have also been exploring various ways of fine-tuning {the power of LLMs} as well as edit-based approaches such as \texttt{GECToR} \citep{omelianchuk-etal-2020-gector} and \texttt{T5} \citep{rothe-etal-2021-simple}, {which currently have provided SOTA results for the GEC task. These pre-trained models are prepared for seamless integration into our system. 
While prompting GPT for the GEC task has also been been investigated, it could not obtain the state of the art results for the GEC task, and this is still prevalent in recent previous work \citep{loem-etal-2023-exploring}.}

\section{Discussion and Future Perspectives}

\paragraph{Features and metrics}
{In the integration,} we have {examined several} automatic metrics as features {to describe the attributes} of the training dataset.
{Specifically, these attributes} are represented in terms of complexity, fluency, and accuracy features. 
{First,} complexity features use quantitative measures such as the number of words and sentences in the text with their numbers and mean lengths. The length of the written text has been considered an essential feature in the learner corpus mostly for identifying proficiency levels. 
Complexity features can also measure syntactic complexity in L2 writing \citep{polio:1997,ortega:2003,lu:2010}.
 Syntactic complexity measures are also considered, which include Yngve's depth algorithm \citep{yngve:1960}, Frazier's local non-terminal numbers \citep{frazier:1985}, and the D-level scale \citep{rosenberg-abbeduto:1987,covington-EtAl:2006} in L1 writing. 
{Second,} fluency{, as}
the potential of language learners to apply their knowledge of grammar to produce intelligible speech and writing, 
{also} plays an important role in language production. 
{For fluency, we considered} the two metrics defined in \citet{asano-mizumoto-inui:2017:IJCNLP} and \citet{ge-wei-zhou:2018:ACL} as {fluency} features of AWE.
{Third,} accuracy represents the ability to produce correct sentences using correct grammar and vocabulary. However, it requires linguistic annotation such as grammatical error categories and error correction. 
{Currently, the} AWE and GEC models have been trained independently. 
We plan to introduce GEC results as accuracy features for the AWE system in future works.

\paragraph{Multilingual aspects for GEC}
{Current system is demonstrated with GEC results for English, but the GEC model is being extended} to the multilingual environment, including Chinese and Korean, using {multilingual Transformers such as} \texttt{mBART} \citep{liu-etal-2020-multilingual-denoising}. 
For Chinese, our current GEC training set is from NLPTEA 2014-2016 CGED Shared Tasks \citep{lee-etal-2016-overview}.\footnote{\url{https://sites.google.com/view/nlptea2018/shared-task}}
It contains 587,050 sentence pairs from writing samples of Chinese L2 language learners (\textit{i.e.}, the writing section of the computer-based 
\textit{Test of Chinese as a Foreign Language} (TOCFL) and of the \textit{Hanyu Shuiping Kaoshi} (HSK) (`Chinese Proficiency Test')). 
Chinese GEC sentences are segmented {into} words using \texttt{pkuseg} in SpaCy\footnote{\url{https://github.com/explosion/spacy-pkuseg}}.
For Korean, the grammatical error correction dataset from National Institute of Korean Language\footnote{\url{https://www.korean.go.kr}} contains 57,320 sentence pairs for training from written writing samples of Korean L2 language learners.
Korean GEC sentences are preprocessed into the sequence of morphemes as a de facto format for Korean machine translation.
With ERRANT \citep{bryant-felice-briscoe:2017:ACL} being adapted to Chinese \citep{hinson-etal-2020-heterogeneous}, Korean \citep{yoon-etal-2023-towards}, and many other languages, it is now possible to streamline grammatical error annotation and evaluation across languages.

\paragraph{Multilingual aspects for AWE and universal rubrics}
Similar to the multilingual GEC, we also adopt a multilingual perspective for our AWE system. Although the current AWE is demonstrated to assess English inputs, it is designed to be compatible with multilingual input data.
This is made possible by two configurations in the assessment process: (1) having the same underlying scoring scheme for the AWE training data regardless of what language is being used, and (2) the set of evaluation standards designed in the scoring scheme assesses the communicative effectiveness of the writers' compositions, rather than any language-specific traits. 
{Both the multilingual scoring scheme and the evaluation standards are ascertained from comparing the proposed language-specific AWE rubrics in the literature, and then extracting the shared analytic traits to serve as the basis for multilingual AWE's training dataset evaluation rubrics}. 
Following the same mechanism, our AWE has the potential to easily expand to incorporate a new language if that language's proposed AWE evaluation rubrics can be accommodated into our existing ones.
In fact, for the Korean AWE system, \citet{lim-song-park-2023} already implemented a holistic scoring system based on the learner corpus from L2 learners of Korean \citep{park-lee:2016}. 
We are currently developing a multilingual rubric scoring system based on the evaluation rubrics from ASAP++ \citep{mathias-bhattacharyya:2018:LREC} for English, ACEA dataset's multi-traits \citep{he-etal-2022-automated} for Chinese, and the rubrics used in the recently released Korean essay evaluation dataset for Korean.\footnote{Released in October 2022, \url{https://aihub.or.kr/aihubdata/data/view.do?currMenu=115&topMenu=100&aihubDataSe=realm&dataSetSn=545}}

\begin{table*}[!ht]
\centering
\scriptsize
{
\begin{tabularx}{\textwidth}{|c X |} \hline 
\multicolumn{2}{|l|}{Perceived fairness and accuracy of GEC and AWE:} \\
&- On a scale of 1-5, how fair do you  think the grammar feedback on this writing to be? (1= not fair at all, 5= very fair)\\
&- On a scale of 1-5, how accurate and trustworthy do you think the grammar correction on this writing to be? (1= not accurate/trustworthy at all, 5= very accurate/trustworthy) \\
&- On a scale of 1-5, how fair do you  think the grading and evaluation of this writing to be? (1= not fair at all, 5= very fair) \\
&- On a scale of 1-5, how accurate and trustworthy do you think the grading and evaluation of this writing to be? (1= not accurate/trustworthy at all, 5= very accurate/trustworthy) \\ \hdashline
\multicolumn{2}{|l|}{Perceived clarity and helpfulness:} \\
&- On a scale of 1-5, how clear do you think the grading and feedback on this writing to be? (1= not clear at all, 5= very clear)\\
&- On a scale of 1-5, how helpful do you perceive the feedback on this writing to be in improving this student's writing skills? (1= not helpful at all, 5= very helpful)\\
&- On a scale of 1-5, how helpful do you think rubrics with criteria and scores to be in understanding the grading and feedback on this writing assignment? (1= not helpful at all, 5= very helpful)\\ \hdashline
\multicolumn{2}{|l|}{Mental effort:}\\
&- On a scale of 1-5, how difficult do you find it to perceive the grading and feedback on this writing to be?\\
&- On a scale of 1-5, how much mental effort does this task (differ for stages 1 and 2) require from you to perceive the feedback?\\ \hdashline
\multicolumn{2}{|l|}{Attitudinal responses:}\\
&- On a scale of 1-5, how comfortable do you feel about perceiving/using this system to receive the grading and feedback on this writing? \\
&- On a scale of 1-5, how effective do you feel about perceiving/using this system to receive the grading and feedback on this writing?\\ \hline 
\end{tabularx}}
    \caption{Questionnaire for the perception experiment of the functionality (GEC+AWE) and effectiveness of the current system}
    \label{experiment-design}
\end{table*}

\begin{figure}[!ht]
\centering
\resizebox{.4\textwidth}{!}{  
\includegraphics{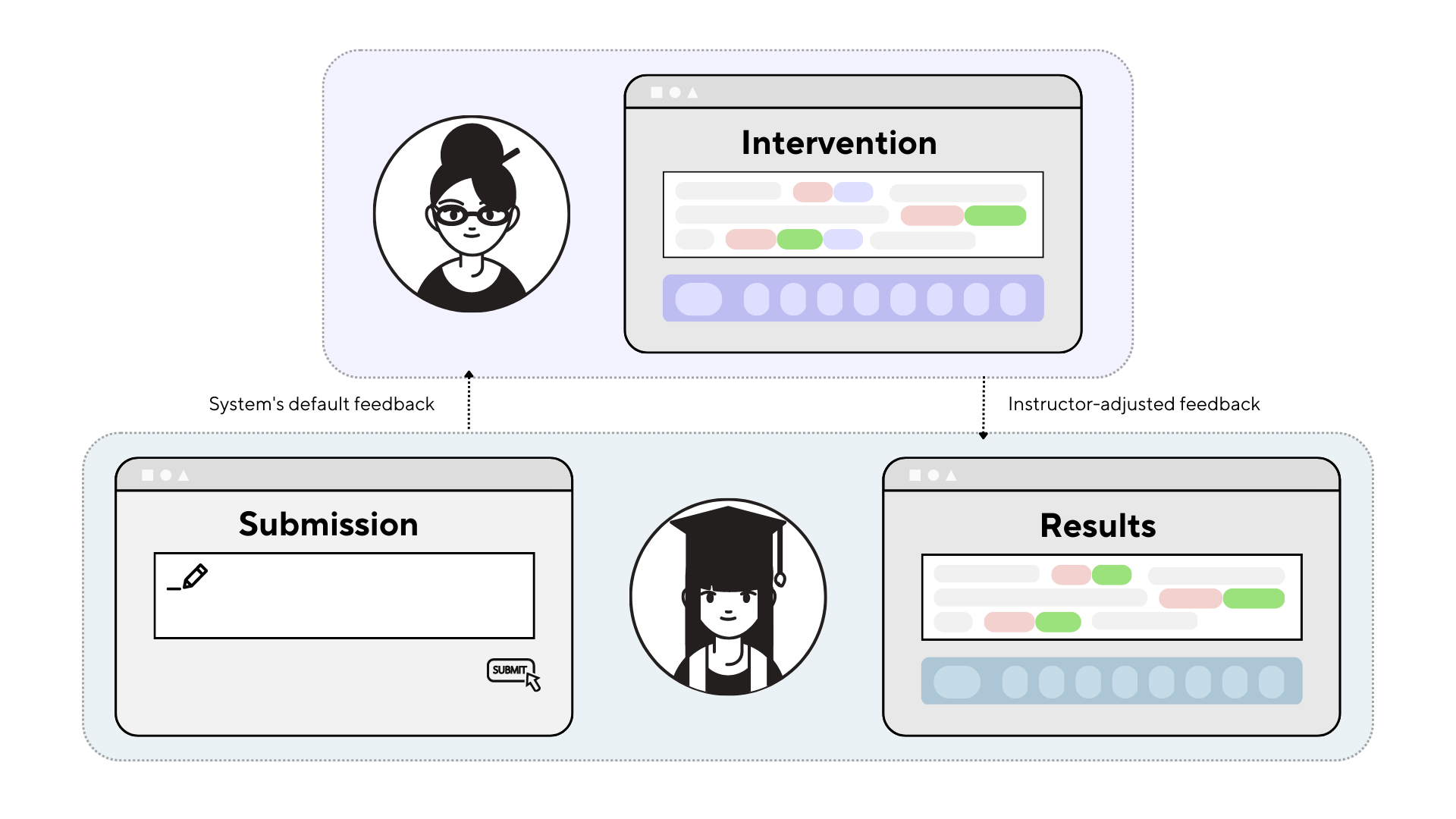}
}
\caption{{Future} system workflow for the language learners and the instructors}
\label{workflow}
\end{figure}

\paragraph{Teacher's view}
The workflow in Figure~\ref{flowchart} focuses on the perspectives of language learners, in which they can prepare for language examinations in a more tangible and efficient way. By simply submitting their complete writing to the system, not only does this workflow enable simulated examination situations, where the language learners receive instant objective scoring results, but also corrective feedback is provided for each submission. 
In addition to serving language learners, Figure~\ref{workflow} indicates another scope, where the system also facilitates language instructors. In this workflow, instructors can intervene with the AWE and GEC process in correspondence with desired rubrics and conventions before the results are delivered to students. This seemingly minor inclusion of the instructor's view capacitates the system's potentiality to extend from simulating examinations for students' practice to facilitating actual examination situations.
Therefore, this integrated system would also alleviate the burden of manually correcting innumerable writings from language instructors. This would enable the instructors to focus their expertise on teaching, thereby providing students with more valuable language learning experiences.

\paragraph{Perception problem}
There are several studies which investigated the effectiveness of AWE on improving learners’ writing performances \citep{chodorow-gamon-tetreault:2010,lee:2020} and students’ perceptions towards AWE tools \citep{chen-cheng:2008,link-EtAl:2014,ranalli:2018,zhang-hyland:2018,oneill-russell:2019,zhang:2020,ariyanto-mukminatien-tresnadewi:2021}.
The majority of {the} research{es} looked into the use of AWE feedback for L2 writing on single-group designs \citep{dikli-bleyle:2014,lavolette-polio-kahng:2015}. Thus, it remains unclear to what extent automated feedback provided by the AWE system supports second language acquisition at various proficiency levels. The present study 
{further investigates the impact of} AWE feedback 
{on} L2 writing performance and students’ perceptions at varying proficiency levels. 
Therefore, we are planning to conduct the following experiments: (i) student perception experiments{:} how language learners perceive results from the automated system without human instructor intervention; (ii) instructor perception experiments: how language instructors perceive and evaluate results from the automated system being conveyed to language learners. 
{,} in collaboration with Centre for Teaching, Learning and Technology at The University of British Columbia 
(See Table~\ref{experiment-design}).\footnote{\url{https://ctlt.ubc.ca}}

\section{Conclusion}

{With the seamless integration of neural AWE and GEC, t}he proposed system will be {made} available to the general public {for the potential to serve as an invaluable language resource for language education.}
{With features that extend from simulating examination settings, this integrated system empowers language learners to improve their writing skills and language proficiency through hands-on mistakes and encourages them to "write it right" to facilitate actual examination situations.
This multifaceted tool offers comprehensive approaches that not only promote better learning outcomes but also transform the traditional interaction between language learners and instructors in language education. These innovative approaches make the proposed system a promising language resource for anyone who seeks to improve the L2 learning experience.}
The proposed system is accessible {at} \url{https://open-writing-evaluation.github.io/awe/}.

\section*{Ethics Statement}
To confirm Behavioural Research Ethics at the University of British Columbia,\footnote{\url{https://ethics.research.ubc.ca/behavioural-research-ethics}} 
authors have obtained a certificate of the Tri-Council Policy Statement: Ethical Conduct for Research Involving Humans (TCPS 2): Course on Research Ethics (CORE-2022).\footnote{\url{https://tcps2core.ca}}

\section*{Acknowledgement}
This research is based upon work partially supported by the \textit{SoTL Seed Program} for Jungyeul Park at The University of British Columbia.


\end{document}